%% file: main.tex
\documentclass[sigconf,nonacm]{acmart}
\usepackage{xspace}
\usepackage{url}
\usepackage{subfigure}
\usepackage{graphicx}

\newcommand{\etal}{~\textit{et al.}}
\newcommand{\Name}{\texttt{ContrastEgo}\xspace} 
\newcommand{\Variant}[1]{ContrastEgo-#1} %placeholder for system name
\newcommand{\methodname}[1]{\textbf{\texttt{#1}}\xspace} %placeholder for system name

\begin{document}

\title{Dynamic Graph Representation Learning for Depression Screening with Transformer}

\input{src/affiliation}

\input{src/abstract}


%%
%% Keywords. The author(s) should pick words that accurately describe
%% the work being presented. Separate the keywords with commas.

\maketitle

\input{src/introduction}

\input{src/preliminaries}
\input{src/overview}
\input{src/methodology}
\input{src/experiment}
\input{src/ablation}

\input{src/relatedwork}

\input{src/conclusion}
\bibliographystyle{ACM-Reference-Format}
\bibliography{citation}
\end{document}

%% file: src/affiliation.tex
\author{Ai-Te Kuo}
\email{aitekuo@auburn.edu}
\affiliation{%
  \institution{Auburn University}
  %\city{Auburn}
  \state{Alabama}
  \country{USA}
  \postcode{36849}
}

\author{Haiquan~Chen}
\email{haiquan.chen@csus.edu}
\affiliation{%
  \institution{California State University, Sacramento}
  \city{}
  \state{California}
  \country{USA}
  \postcode{95819}
}

\author{Yu-Hsuan Kuo}
\authornote{Work was done outside Amazon.} 
\email{yuhsuank@amazon.com}
\affiliation{%
  \institution{Amazon}
%  \city{State College}
  \state{California}
  \country{USA}
  \postcode{}
}
\author{Wei-Shinn Ku}
\email{weishinn@auburn.edu}
\affiliation{%
  \institution{Auburn University}
  %\city{Auburn}
  \state{Alabama}
  \country{USA}
  \postcode{36849}
}

%% file: src/abstract.tex
\begin{abstract}

Early detection of mental disorder is crucial as it enables prompt intervention and treatment, which can greatly improve outcomes for individuals suffering from debilitating mental affliction. The recent proliferation of mental health discussions on social media platforms presents research opportunities to investigate mental health and potentially detect instances of mental illness. However, existing depression detection methods are constrained due to two major limitations: (1) the reliance on feature engineering and (2) the lack of consideration for time-varying factors. Specifically, these methods require extensive feature engineering and domain knowledge, which heavily rely on the amount, quality, and type of user-generated content. Moreover, these methods ignore the important impact of time-varying factors on depression detection, such as the dynamics of linguistic patterns and interpersonal interactive behaviors over time on social media (e.g., replies, mentions, and quote-tweets).

To tackle these limitations, we propose an early depression detection framework, \Name, that effectively learns multi-scale user latent representations (node-level and graph-level) by capturing the topological dynamics of users' social interactions with their friends using attention mechanism.
Specifically, \Name treats each user as a dynamic time-evolving attributed graph (ego-network) and leverages supervised contrastive learning to maximize the agreement of users' representations at different scales while minimizing the agreement of users' representations to differentiate between depressed and control groups. \Name embraces four modules, (1) constructing users' heterogeneous interactive graphs, (2) extracting the representations of users' interaction snapshots using graph neural networks, (3) modeling the sequences of snapshots using attention mechanism, and (4) depression detection using contrastive learning. Extensive experiments on Twitter data demonstrate that \Name significantly outperforms the state-of-the-art methods in terms
of all the effectiveness metrics in various experimental settings. Compared to all baseline methods,
\Name is able to effectively predict early depression in Twitter
users with up to 5\% and 4\% improvement on F1 score and AUROC, respectively.

\end{abstract}

%% file: src/introduction.tex
\section{Introduction} 
Depression is a prevalent and pernicious global health issue that affects millions of people worldwide. The World Health Organization's report~\cite{who} stated that the cost to the global economy by depression and anxiety disorders was estimated to be one trillion dollars annually, primarily due to diminished productivity. This loss not only affects individuals and families, but also has far-reaching implications for society as a whole. Left untreated, depression can cause significant distress and impairments, leading to a range of severe complications such as suicide, substance abuse, absenteeism, reduced productivity, and other mental health disorders. Early detection and treatment of depression are crucial in avoiding the exacerbation of the condition. The  prolongation of untreated depression can result in a manifestation that becomes increasingly severe and resistant. Despite the significance of early detection and treatment, fear of mental illness stigma often impedes individuals from seeking help, even in the face of debilitating symptoms, presenting a significant barrier to care. Consequently, identifying individuals at risk of developing depression and providing them with early intervention and support becomes more challenging. Fortunately, there has been a recent surge in individuals sharing their mental health experiences and openly discussing their mental health issues on social media platforms. This presents a valuable opportunity for researchers to gain insights into the mental health and well-being of individuals by leveraging user-generated content shared on these platforms. The effective analysis of social media data holds the potential to lead to early detection of individuals who may be at risk of developing depression.

\vspace{1mm}
\noindent\textbf{Existing approaches and limitations.}
Studies in the detection of depression in social media are predominantly focused on the exploitation of text-based features. Several classical machine learning classification models have been utilized to predict depression from social media data, exhibiting reasonable accuracy. These methods include Naive Bayes Classifier~\cite{Chatterjee_2021,8389299}, Support Vector Machine~\cite{DeChoudhury_Gamon_Counts_Horvitz_2021}, and Logistic Regression~\cite{ijcai2017p536, preotiuc-pietro-etal-2015-role,pmid30344616}. However, these approaches may be impeded by the need for copious amounts of data to train, and their ability to process large amounts of data may be inferior to that of deep learning-based methods. In contrast, deep learning-based techniques for depression detection leverage the advanced representation-learning capabilities of neural networks. These methods have been applied to capture complex patterns and relationships in social media data, including text, images, and social network structures, to predict depression. These techniques include using Convolutional Neural Networks (CNNs)~\cite{yates-etal-2017-depression, husseini-orabi-etal-2018-deep}, Recurrent Neural Networks (RNNs)~\cite{husseini-orabi-etal-2018-deep}, Long Short-Term Memory (LSTM) networks~\cite{9447025, 9231008}, Gated Recurrent Unit (GRU) networks~\cite{10.1145/3404835.3462938}, and Graph Neural Networks (GNNs)~\cite{mentalnet}. Prior studies ~\cite{9447025,9231008, husseini-orabi-etal-2018-deep, 10.1145/3459637.3482366, Ophir2020} have focused on the application of language or sentiment analysis to identify words or phrases commonly associated with depression, such as the presence of negative words or terms such as "sad," "hopeless," and "worthless." However, these prediction performances are highly contingent upon the quality and quantity of the data used, with explicit language usage being easier to predict than more subtle or indirect language usage. Additionally, collecting sufficient depression-related data to train domain-specific  word embeddings can be  challenging. Some recent studies have opted to leverage pretrained models~\cite{FIGUEREDO2022100225,s-antony-2022-ssn} for the detection of depression. Recent studies~\cite{10.1145/3459637.3482366, mentalnet,DeChoudhury_Gamon_Counts_Horvitz_2021} have also emphasized the significance of social circles and their impacts on an individual's mental state, highlighting the need for a more holistic
approach to depression detection in social media.

\vspace{1mm}
\noindent\textbf{Our observations and contributions.} We make two important observations.  First, \textit{the inability of existing solutions
to support early depression detection is a serious limitation in practice}. Existing depression detection methods  heavily rely on the amount, quality, and type of user-generated content on social media, thus failing to deal with those depressed users with only limited social footprints.  As a result, those methods tend to suffer from the cold-start problem and fail to detect depression effectively at an early stage.  However, the early detection of depression plays a vital role in supporting
healthcare professionals for determining timely and targeted interventions
in a clinical setting.

Second, \textit{the prediction of the onset of psychiatric diseases (including mental depression) may benefit from treating each individual as a dynamic system with time-evolving structure and attributes (e.g., a dynamic graph consisting of multiple snapshots of clinical states)}.   However, the existing depression detection methods fail to take into account the important
impact of the time-varying factors on depression detection, such as the dynamics of the linguistic patterns and interpersonal interactive behaviors over
time on social media. 

%For instance, an individual in the early stages of depression may exhibit changes in social behavior patterns. These studies typically rely on the most recent user-generated content around the time of diagnosis. However, individuals with depression may be more likely to post negative content in the lead-up to their diagnosis. Moreover, collecting excessive data may be unrealistic, while negative sentiments expressed may be too dominant to be easily detected by modern NLP models. Thus, we seek not only to utilize the dynamic changes in an individual's behavior and social influence over time but also to handle scenarios where limited information is available about an individual, e.g., cold-start problems. The overall objective is to detect patterns within an individual's social network that may have the potential to result in depression, with a limited quantity of data in each period.

%We are impelled to explore an alternative approach to the prevalent use of feature engineering in depression detection, which demands in-depth domain knowledge to extract information pertaining to depression from raw data. The recent breakthroughs in human language understanding make the use of pre-trained NLP models a compelling option, given the informal nature of the language used on social media. 

In this paper, we propose an early depression detection framework, \Name, that effectively learns multi-scale user latent representations (node-level and graph-level) by capturing the topological dynamics of users' social interactions with their friends using attention mechanism.
Specifically, \Name treats each user as a dynamic time-evolving attributed graph (ego-network) and leverages supervised contrastive learning to differentiate between depressed and control groups. Our contributions can be summarized as follows:

%This is difficult because, in a static graph, the heterogeneity of interactions in a static graph is easily accommodated throughout history, but imposing such heterogeneity within a limited time frame in ego networks poses a formidable challenge. If any type of interaction is missing in a period, the graph convolution will then operate on trivial graphs. Although this issue can be offset by imposing a strong constraint requiring every type of interaction in every period during data preparation, it will then mean that our approach is only applicable to a small group of users who spend a significant amount of time posting content and interacting with other users. The other issue is that with the learned representations of a user in different time periods, how do we know to which part of the information it needs to pay attention when making predictions? Additionally, current solutions of representation learning on temporal graphs ingest sequences in a sequential manner, which can make it more difficult to parallelize. Motivated by these limitations, this paper proposes a framework for learning on temporal graphs for the early detection of depression. %To the best of our knowledge, our work is the first to temporally model the users’ social circles.

\begin{enumerate}
    \item \Name treats the dynamic nature of user interactions on social media as first-class citizens to detect depression at an early stage. To the best of our knowledge, \Name is among the first to consider the early detection of depression with temporal aspects. Our research findings emphasize the significance of pattern changes inherent in mental disorders and provide a novel and insightful viewpoint. This could result in a more efficient and effective early detection for individuals who may be at risk of mental illness for timely and targeted interventions.
    
    %\item We alleviate the reliance on feature engineering for detection of depression by utilizing robust pre-trained NLP models, specifically trained on social media content. \yk{[we do not "alleviate the reliance on feature engineering". we do not use feature engineering.]}
    \item  In order to capture the interpersonal dynamics on social media to identify depression status, \Name utilizes graph neural networks to encode social interactions and extract the latent user representations for each snapshot at multiple scales (at the node level and the graph level). 

    \item In order to capture the temporal dynamics
    (e.g., abrupt versus gradual  onset) of depression,  \Name utilizes a transformer architecture to encode the sequences of snapshots using attention mechanism.

    \item \Name adopts supervised contrastive learning~\cite{NEURIPS2020_d89a66c7} to maximize the agreement of users’ representations extracted at different scales and minimize the agreement of users’ representations between the individuals with depression and those without depression.

    \item Extensive experiments on real-world Twitter data verify that \Name consistently outperforms the state-of-the-art methods in various experimental settings. In particular, \Name has shown an improvement of up to 4\% in the AUROC score for predicting early depression in Twitter users compared to the state-of-the-art methods.
\end{enumerate}

The organization of this paper is as follows. In Section~\ref{sec:prelim}, we present background information. In Section~\ref{sec:overview}, we provide an overview of \Name. In  Section~\ref{sec:methodologies}, we explain the building blocks of \Name. Section~\ref{sec:exp} compares the performance of \Name with that of the state-of-the-art methods. Section~\ref{sec:related_work} reviews the related work while Section~\ref{sec:conclusion} concludes the paper.

%% file: src/preliminaries.tex
\section{Preliminaries}\label{sec:prelim}

\noindent\textbf{Graph snapshot.}
A graph snapshot $G_t = (V_t, E_t)$ is a representation of a graph at a specific point in time $t$, where $V$ is a set of vertices and $E$ is a set of edges connecting the vertices. In our model, a graph snapshot of a user $U$ represents a collection of interactions around $U$ within a specific time frame. The vertices in the graph $V$ correspond to the individuals involved in the interactions, and the edges $E$ represent the interactions between individuals and their friends. Each vertex is represented by a vector of tweet embeddings, and each edge represents a type of interaction and the time period of the interaction.

\vspace{1mm}
\noindent\textbf{Dynamic Graph.} A dynamic graph can be formally defined as a sequence of graph snapshots with changing topological structures $G_1, G_2, ..., G_T$, where each snapshot $G_t = (V_t, E_t)$ represents the state of the graph at time $t$. Here, $V_t$ is the set of vertices and $E_t$ is the set of edges in the graph at time $t$. The vertices and edges in a dynamic graph may vary over time, with new vertices and edges being added or removed as the graph evolves. In addition, the attributes of the vertices and edges may also change over time. For instance, consider a dynamic social media graph representing interactions between users. The vertices in the graph represent the users, and the edges represent the interactions between users, such as replies to a post or mention in a comment. The attributes of a vertex, represented by tweet embeddings, may change over time as the user posts new comments.  In addition, edges may appear or disappear in the graph as the connections between users change over time. For example, an edge may be added to the graph if two users begin interacting with each other, and it may be removed if the interaction ceases.

\vspace{1mm}
\noindent\textbf{Egocentric Network.} Ego networks are a type of graph structure widely used in social network analysis. An ego network consists of an ego node (central node) and the relationships between the ego node (user) and the alter nodes (friends) in the network. 

%% file: src/overview.tex
\section{Overview}\label{sec:overview}

\subsection{Problem Definition}
Given a dataset $D = \{{d_{1}, \dots, d_{n}}\}$, each individual $d_i$ is represented by a sequence of egocentric networks over $T$ time step, each of which contains social interactions on social media in an observation period, where node features on each graph are represented by their tweets during the respective period. The problem is to predict, for each individual $d_i$, whether they have depression or not.

\begin{figure*}[!htbp]
    \centering
    \includegraphics[width=1.0\linewidth]{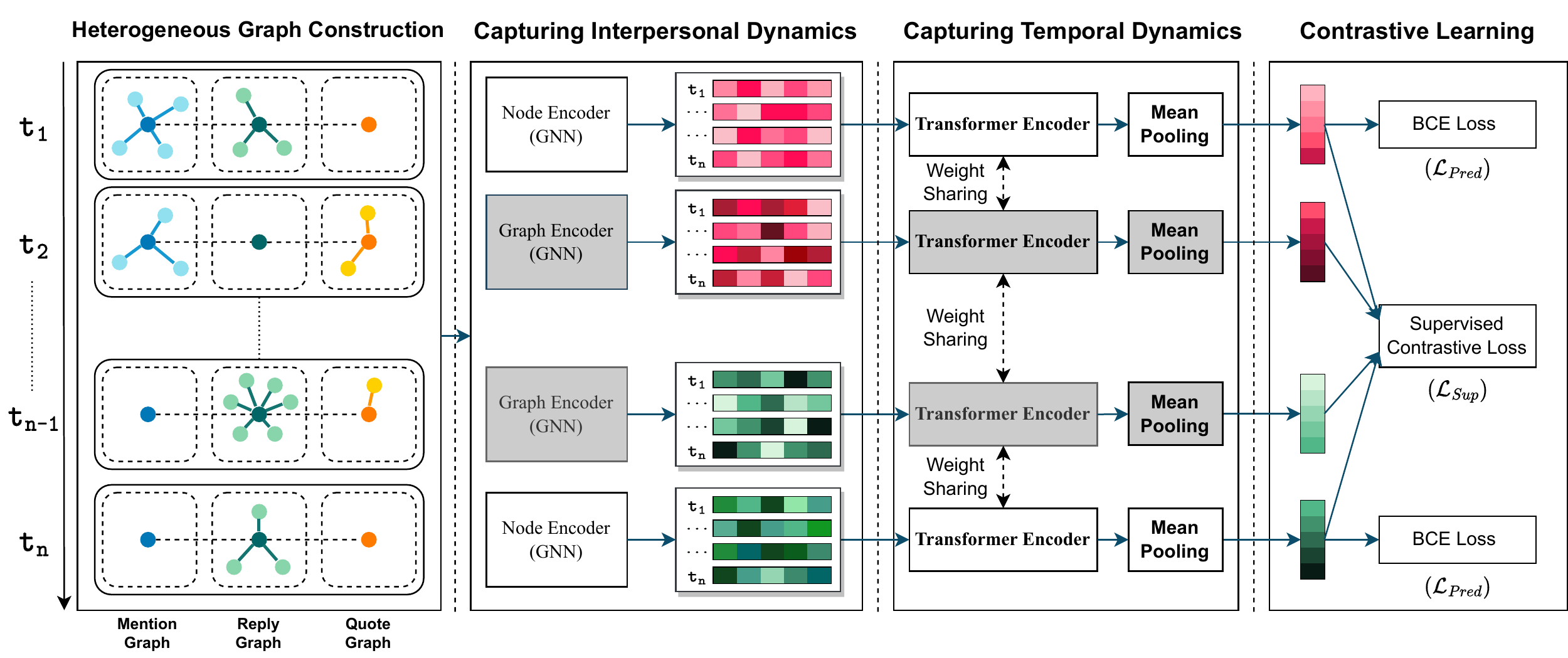}
    \caption{Architecture of \Name}
  \label{figure:arch}
\end{figure*}

\subsection{\Name architecture.} \Name embraces four modules, (1) constructing users' heterogeneous interactive graphs, (2) extracting the representations of users' interaction snapshots using graph neural networks, (3) modeling the sequences of snapshots using attention mechanism, and (4) depression detection using contrastive learning. We first use a pre-trained RoBERTa model~\cite{loureiro-etal-2022-timelms, barbieri-etal-2020-tweeteval}, trained on the Twitter corpus, to obtain the embeddings of all tweets posted by each user during each time period (snapshot).  The averaged embedding obtained for each user is treated as the attributes of the corresponding node in each user's ego-network. 
Next, for each user, we construct her heterogeneous interactive graph, where each snapshot in such dynamic graph contains the three distinct types of interactions between users and their friends (i.e., replies, mentions, and quote-tweets). In order to capture the interpersonal dynamics on social
media to identify depression status for each snapshot, we utilize
graph neural networks to encode social interactions and
extract the latent user representations for each snapshot at
multiple scales. In order to capture the temporal dynamics of depression, we utilize a transformer architecture to encode the sequences of snapshots using attention mechanism. At last, we adopt supervised contrastive learning to
maximize the agreement of users’ representations extracted
at different scales and minimize the agreement of users’
representations between the individuals with depression
and those without depression.
Given the contrastive loss $\mathcal{L}_{Con}$ and the binary cross-entropy loss $\mathcal{L}_{Pred}$, the loss function which \Name aims to optimize 
can be represented as follows:

\begin{equation}
    \mathcal{L} =  \alpha \mathcal{L}_{Con} + \mathcal{L}_{Pred},
\end{equation}
where $\alpha$ is a trade-off coefficient. 
%\yk{[add $\beta$ in front of $L_{pred}$?]}

%% file: src/methodology.tex
\section{Methodologies}\label{sec:methodologies}
\subsection{Heterogeneous interactive graphs}
Feature engineering, a prevalent approach in prior literature~\cite{Chatterjee_2021,8389299,preotiuc-pietro-etal-2015-role, DeChoudhury_Gamon_Counts_Horvitz_2021,8681445,ijcai2017p536,9447025}, has been known to have many limitations. To capture the linguistic characteristics of users and their friends, \Name utilizes pre-trained NLP models such as BERT \cite{devlin2018bert}, XLNet \cite{yang2019xlnet} and RoBERTa \cite{liu2019roberta} to extract  the embeddings of all tweets posted by each user/friend during each time period (snapshot). The averaged embedding obtained for each user/friend is treated as the attributes of the corresponding
node in each user’s ego-network.

For each user, we create her ego-network graph for each specific interaction
type where each edge attribute value is set as the number of interactions
in that type between the user and each of her friends. The embeddings obtained
from the pretrained RoBERTa model are adopted as the node features. We then merge these three ego-network graphs for each user into a heterogeneous graph with
each edge being a 3-dimensional vector representing the strengths of different interaction types (replies, mentions, or quote-tweets).
Given a user’s information, we only keep those friends with whom she interacts. In other words, if a friend is missing all three interaction types with the associated user, that friend is pruned. By constructing a heterogeneous graph to model the interactions within the graph, we aim to better capture the complexity and richness of real-world social networks and gain insights into how social interactions can influence behavior.

In addition, we also propose a graph combination technique to enhance training efficiency. Instead of processing each time period one at a time, we combine all the graphs together for efficient computation. Let $G_t = (V_t, E_t)$ denote the snapshot of the graph at time step $t$, where $|V_t|$ represents the cardinality of nodes and $|E_t|$ represents the cardinality of edges. The aggregated graph $G' = (V', E')$ is defined as the union of all graph snapshots from an individual, such that $V' = \bigcup_{t=1}^{T} V_t$ and $E' = \bigcup_{t=1}^{T} E_t$, where $T$ represents the total number of time periods. The resulting graph $G'$ provides a global perspective of the ego nodes over time, where each graph snapshot is considered as a connected component. The properties of the resulting graph $G'$ are: (1) $|V'| = \sum_{t=1}^{T} |V_{t}|$, (2) $|E'| = \sum_{t=1}^{T} |E_{t}|$, and (3) the number of connected components of $G'$ equals $T$. This implies that the nodes in G' represent the same entities at different points in time, but they are not directly connected to each other in the aggregate graph. This aggregated graph is then used for capturing interpersonal dynamics.

%The tweet embeddings generated by the pre-trained model can be mathematically represented as a $d$-dimensional vector, where $d$ is the dimension of the embeddings. Specifically, for a tweet $T$, tweet embeddings $E$ can be represented as $E = f(T; \theta)$, where $f(T; \theta)$ is the function that maps the tweet $T$ to the $d$-dimensional embeddings E, and $\theta$ are the fixed, learned parameters of the pre-trained NLP model. The utilization of pre-trained NLP models to generate tweet embeddings enables the extraction of highly informative features, such as sentiment, topic, and emotion, from tweets. These embeddings, encapsulating the meaning of tweets in a concise and informative manner, can be fed into a GNN to learn representations that capture the interactions between users and friends in the graph. 

%By treating these embeddings as node features, our downstream task of early detection of depression can greatly benefit from their ability to extract valuable information. By leveraging these generated tweet embeddings as node features, our proposed approach for early detection of depression can benefit from their ability to extract valuable information. Furthermore, by viewing the generated embeddings as the progression of the user's behavior, we can gain insight into subtle variations and changes in their linguistic patterns, emotions, and social interactions. This can provide a more comprehensive and accurate representation of the user, which can aid in the early detection of depression.

\subsection{Capturing interpersonal dynamics}
Recently, several studies have been proposed to extract multi-scale substructure features that are critical in understanding the underlying relationships in social networks using graph convolution~\cite{gcnReview,GCN}.
Let $G$ be an undirected graph of $n \in \mathbb{N}$ nodes and an adjacency matrix $\mathbf{A} \in \mathbb{Z}^{n \times n}$. An augmented adjacency matrix of $G$ is defined as $\mathbf{\tilde{A}}=\mathbf{A}+I_{n}$. For row-wise normalization, a diagonal degree matrix of $\mathbf{D}$ is defined as $\mathbf{\tilde{D}}_{i, i}=\sum_{j} \mathbf{\tilde{A}}_{i, j}$. A matrix of learnable parameters is defined as $\mathbf{W} \in \mathbb{R}^{c \times c^{\prime}}, c^{\prime} \in \mathbb{N}$, where $c^{\prime}$ is the number of output feature channels. Suppose that the node feature matrix is $\mathbf{X} \in \mathbb{R}^{n \times c}, c \in \mathbb{N}$ and a non-linear activation function is denoted by $\sigma$. The graph convolution operation can be defined as follows:
\begin{equation}
\mathbf{Z}=\sigma\left(\tilde{\mathbf{D}}^{-1} \tilde{\mathbf{A}} \mathbf{X} \mathbf{W}\right),
\label{eq_dgcnn}
\end{equation}
where graph convolution aggregates information about the local substructure by smoothing information about nodes in the local neighborhood. To extract the multi-scale substructure features, multiple graph convolution layers are stacked, as follows:
\begin{equation}
\mathbf{Z}^{(l+1)}=\sigma\left(\tilde{\mathbf{D}}^{-1} \tilde{\mathbf{A}} \mathbf{Z}^{(l)} \mathbf{W}^{(l)}\right),
\label{eq_dgcnn_multi}
\end{equation}
where $\mathbf{Z}^{(0)}=X$ and $\mathbf{W}^{(l)} \in \mathbb{R}^{c_{l} \times c_{l+1}}$.

\Name adopts graph convolution  to capture the interpersonal interactions within each snapshot in the created dynamic heterogeneous interactive graph. \Name updates the node embeddings represented by $H^{(l+1)}$ as follows:

\begin{equation}
\begin{split}
    \mathbf{H}^{(l+1)} \leftarrow \sigma(\sum_{r \in R}(\tilde{\mathbf{A}_{r}}\mathbf{H}^{(l)}\mathbf{W}^{(l)}_{r})),
\end{split}
\end{equation}

\noindent where $\mathbf{H}^{(l)}$ refers to the node features when $l=0$, and the activation output of the $l^{th}$ layer for $l > 0$, $\tilde{\mathbf{A}_{r}}$ is the normalized adjacency matrix for edge type $r$, $\mathbf{W}^{(l)}_{r}$ is the learnable weight matrix for layer $l$ with respect to edge type $r$, $R$ represents a set of edge types (mention, reply, and quote). The GELU activation function was used in our experiments.

\subsection{Capturing temporal dynamics}

Attention mechanism~\cite{NIPS2017_3f5ee243} has greatly improved the ability to capture dependencies among the elements of sequential input, especially in long sequences. \Name leverages a transformer encoder, which comprises multi-head self-attention networks and feed-forward neural networks to capture the temporal dynamics. The input sequences for the transformer encoder are created by rearranging the outputs of the heterogeneous graph learning module. The transformer encoder operates by computing three distinct representations of the input sequence. These representations, namely the query $Q = XW_{Q}$, key $K = XW_{K}$, and value $V = XW_{V}$ representations, are computed by applying linear transformations to the input sequence $X$, using learnable parameters $W_Q$, $W_K$, and $W_V$, respectively. Subsequently, attention scores for a sequence are computed as follows: \begin{equation}
     Attention(Q, K, V) = softmax(\frac{QK^T}{\sqrt{d_{k}}})V
 \end{equation}
 \noindent where $d_k$ is the dimension of $K$. The attention scores represent varying levels of importance to the ego nodes at different time periods, based on their relevance to the current query representation. The final output representation of the sequence is then derived by taking the dot-product of the attention scores and value representation. To ensure accurate evaluations of depression progression, we adopted positional encodings~\cite{NIPS2017_3f5ee243} into the transformer encoder. This enhances the model's perception of the relative positions of ego nodes in the input sequence and enables it to understand temporal relationships between changes in a person's social media behavior, such as post interactions and content. Proper consideration of the order of these behaviors is crucial in determining depression progression, as the arrangement can significantly impact the interpretation of a person's mental state. For example, a decrease in depression symptoms followed by an increase may indicate worsening, while the reverse scenario may indicate improvement. The positional encodings, denoted by $\{p^1, \dots, p^{t}\} \in \mathcal{R}^{d}$, are added element-wise to the input sequence $x^i$, where $i \in [1, t]$ and $t$ is the total number of time periods. Mathematically, the element-wise addition can be represented as $x^i + p^i$, where $x^i \in \mathcal{R}^{d}$ is the input embedding of the $i^{th}$ time period and $p^i \in \mathcal{R}^{d}$ is the positional embedding of the $i^{th}$ time period. After a fixed positional encoding vector $p^i \in \mathbb{R}^d$ is added to each input element of the sequence, the resulting sequence is then passed through the transformer encoder. To obtain a comprehensive user representation, we opt for a more simplistic approach by utilizing mean pooling on the encoded sequence rather than introducing a [CLS] token for classification.

\begin{equation}
\begin{split}
    \mathcal{X} &\leftarrow Pooling(TransformerEncoder( \{x^{i} + p^{i}\})), i \in [1,t]\\
\end{split}
\end{equation}

\subsection{Supervised Contrastive Learning}

\Name utilizes a supervised contrastive learning method to maximize the agreement of users’ representations extracted at different scales and minimize the agreement of users’
representations between the individuals with depression and those without depression. As shown in Figure~\ref{figure:arch}, the node encoder and the graph encoder constitute the node- and graph-level views, respectively. Both views are then fed into the same transformer encoder, which learns to attend to temporal changes of the egos as well as the impact of users' social circles. The transformer encoder's output is then further processed by applying mean pooling over each sequence representation to obtain the overall node- and graph-level representations of the user for depression detection. The objective of supervised contrastive learning is to learn the representations of the data such that the representations of the positive pairs are closer to each other and the representations of the negative pairs are farther apart in the embedding space. Thus, we employ a projection neural network for contrastive learning to encourage the network to maximize agreement between positive pairs (same-group node-graph) and disagreement between negative pairs (different-group node-graph). This is predicated on the understanding that a person's social milieu, including their relationships and social support, plays a role in the progression of depression~\cite{10.1145/3459637.3482366, mentalnet,DeChoudhury_Gamon_Counts_Horvitz_2021}. A popular contrastive learning framework, SimCLR~\cite{chen2020simple}, only permits one positive sample to be paired with each anchor and considers all other augmented examples as negative examples. \Name adopts supervised contrastive loss~\cite{NEURIPS2020_d89a66c7}, which extends the self-supervised contrastive loss by taking into account class label information. The supervised contrastive loss is formulated as follows:

\begin{equation}
\mathcal{L}_{sup} = \sum_{u \in U}\frac{-1}{|P(u)|}\sum_{p \in P(u)}\log\frac{e^{(z_{u} \cdot z_{p}/\tau)}}{\sum_{a \in A(u)}e^{(z_{u}\cdot z_{a}/\tau)}}
\end{equation}

\noindent where $U$ is the set of all unique samples in the multiviewed batch, $P(u)$ is the set of indices of all positives in the multiviewed batch distinct from $u$, $z_{u}$ and $z_{p}$ are the embeddings of anchor and positive, respectively and $A(u)$ is the set of all negatives in the multiviewed batch distinct from $u$. Here, $\tau$ is a temperature hyperparameter.
\begin{figure}[!htbp]
    \centering
    \includegraphics[width=0.85\linewidth]{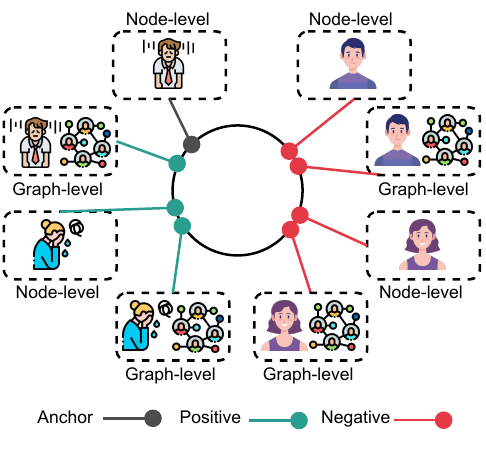}
    \caption{Illustration of Supervised Contrastive Learning}
  \label{figure:supcon}
\end{figure}

Figure~\ref{figure:supcon} visually represents the concept of supervised contrastive learning in \Name. This approach utilizes both node-level and graph-level embeddings to enhance the learning process. To demonstrate this concept, we take the example of an individual with depression as the anchor. In supervised contrastive learning, individuals with depression (node-level view) and their social circles (graph-level view) are considered positive examples while the rest of the control group is regarded as negatives. This approach results in an embedding space where individuals belonging to the same class (depression) are grouped closer together, while individuals from different classes (control group) are separated further apart.

%% file: src/experiment.tex
\section{Experiment}\label{sec:exp}
In this section, we empirically show the superiority of \Name
over the state-of-the-art methods.

\subsection{Baselines}\label{sec:baselines}
\begin{itemize}
    \item \methodname{GCN}~\cite{GCN}. This method utilizes the Laplacian matrix of the graph to perform convolution operations on the node features.
    \item \methodname{GraphSAGE}~\cite{graphsage}. This method uses k-hop sampling to aggregate features from the sampled local neighborhood.
    \item \methodname{GAT}~\cite{gat}. This method employs self-attention mechanisms to weigh the importance of different nodes in the ego's local network.
    \item \methodname{MentalNet}~\cite{mentalnet}. This method models three different types of interaction on social media (reply, mention, and quote), after which DGCNN~\cite{DGCNN} is applied to generate the user representation as a graph. The LSTM autoencoder was replaced by a pretrained RoBERTa model~\cite{loureiro-etal-2022-timelms, barbieri-etal-2020-tweeteval} in our experiments for a fair comparison.
    \item \methodname{DySAT}~\cite{dysat}. This method leverages a GAT layer to obtain structural information and later employs self-attention mechanisms to capture temporal dependencies.
\end{itemize}

\renewcommand{\arraystretch}{1.2}
\begin{table*}[t]
  \caption{Statistics of PsycheNet-G}
%   \footnotesize
  \small
  \centering
  \setlength{\tabcolsep}{0.35em}
  \begin{tabular}{@{}lrrrrrrrrrrrrrrrrrp{1.5cm}p{1.5cm}@{}}
    \toprule
    %\multirow{2}{*}{}
    & \multicolumn{1}{c}{\textbf{Users}} & \multicolumn{4}{c}{\textbf{Friends per user}} & \multicolumn{4}{c}{\textbf{Replies per user}}
    & \multicolumn{4}{c}{\textbf{Mentions per user}} & \multicolumn{4}{c}{\textbf{Quote-Tweets per friend}} \\
    \cmidrule(lr){2-2}  \cmidrule(lr){3-6} \cmidrule(lr){7-10} \cmidrule(lr){11-14} \cmidrule(lr){15-18}
    & \textbf{Sum}
    & \textbf{Min} & \textbf{Max} & \textbf{Mean} & \textbf{Sum}
    & \textbf{Min} & \textbf{Max} & \textbf{Mean} & \textbf{Sum}
    & \textbf{Min} & \textbf{Max} & \textbf{Mean} & \textbf{Sum}
    & \textbf{Min} & \textbf{Max} & \textbf{Mean} & \textbf{Sum} \\
    \midrule
    \textbf{Depressed group} & $242$ & $1$ & $518$ & $63$ & $15,321$ & $1$ & $19,414$ & $1,379$ & $333,799$ & $1$ & $17,885$ & $891$ & $215,722$ & $1$ & $461$ & $35$ & $8,470$ \\
    \textbf{Control group}   & $349$ & $1$ & $448$ & $89$ & $31,103$ & $1$ & $132,907$ & $3,119$ & $1,088,470$ & $1$ & $63,532$ & $1,804$ & $629,437$ & $1$ & $22,328$ & $283$ & $98,700$ \\
    \bottomrule
  \end{tabular}
  \label{tbl:rel_dataset}
\end{table*}
\renewcommand{\arraystretch}{1.1}

\subsection{Dataset}
In our study, we utilized the PsycheNet-G dataset~\cite{mentalnet}, composed of 242  users diagnosed with depression and 349 individuals as the control group. Table~\ref{tbl:rel_dataset} shows the statistics of the PsycheNet-G dataset.
To generate user representations, each tweet was processed to create a tweet embedding utilizing the pre-trained RoBERTa model~\cite{roberta}, \textit{twitter-roberta-base-sep2022}~\cite{loureiro-etal-2022-timelms, barbieri-etal-2020-tweeteval}, specifically trained on millions of tweets. In the case of dynamic/temporal neural networks, the node features of users and friends were represented by the mean of the corresponding tweet embeddings. On the other hand, for non-temporal neural networks, such as \methodname{MentalNet}, \methodname{GCN}, \methodname{GAT}, and \methodname{GraphSAGE}, the node features of users and friends were represented by the mean of the tweet embeddings throughout the user's history, and the adjacent matrix was represented by the aggregation of the adjacent matrices of all time periods. 

\subsection{Research Questions}
\begin{figure}[!htbp]
    \centering
    \includegraphics[width=0.8\linewidth]{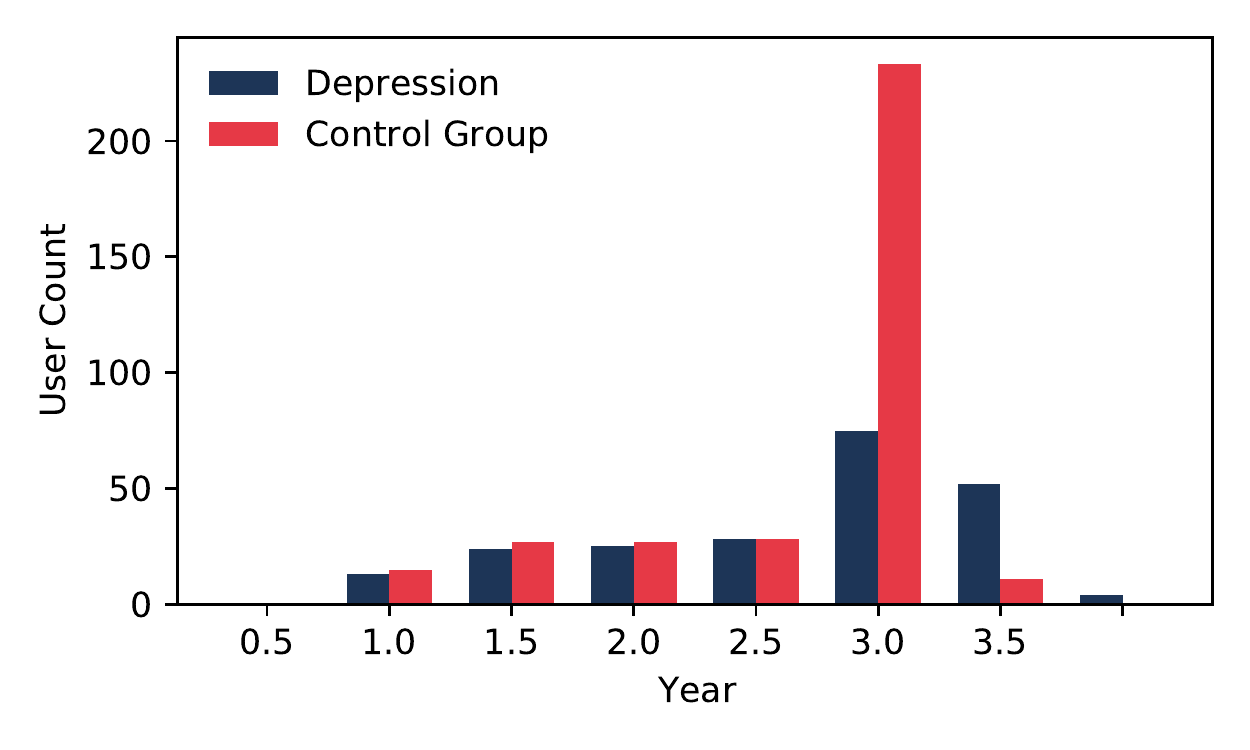}
    \caption{The Distribution of the PsycheNet-G dataset}
  \label{figure:stats}
\end{figure}

To evaluate the effectiveness of our proposed model, we evaluated our model based on three research questions:
\begin{itemize}
    \item [\textbf{(RQ$_1$)}] To what extent does the availability of user-generated content affect the accuracy of depression prediction in a scenario where access and usage of user data are limited?
    \item [\textbf{(RQ$_2$)}] Is incorporating information from an individual's social network an effective method for detecting depression?
    \item [\textbf{(RQ$_3$)}] How does the duration of observation from social networks affect the prediction of depression onset and progression?
\end{itemize}
Based on these research questions, we processed the PsycheNet-G dataset by imposing different numbers of tweets, friends, and periods, as specified in Section~\ref{sec:eval}. For different numbers of tweets per period and the number of friends per period, we applied uniform random sampling of tweets and friends, respectively. Any data in a period where the user had insufficient tweets or friends were not considered. For varying the length of every period, any data in a period where the user had insufficient tweets and friends were not considered. Figure~\ref{figure:stats} illustrates the distribution of data collection duration for all users. The y-axis represents the number of users whose data were collected for a certain duration.

\subsection{Experiment Settings}
%\yk{break into 3 parts/paragraphs: exp. settings and evaluation protocol,  parameters,  how you implemented}
In this study, we presented the performance of each method in terms of precision, recall, F$_1$, and AUC-ROC scores. All experiments were performed using Python v3.9, PyTorch v1.11.0 with an Intel(R) Core(TM) i7-7700HQ CPU (at 2.80GHz) and a GTX 1070Ti GPU (with 8 GB RAM). The model evaluation was performed using rigorous five-fold cross-validation evaluations, and the reported results were the average of ten runs. We obtained the implementation code for the baseline methods either from the PyTorch Geometric package or their official web pages. Unless otherwise specified, the default number of tweets per period is 5, the number of friends per period is 4, and the duration of each period is 12 months. To facilitate reproducibility, Table~\ref{table:hyperparams} presents the hyperparameter configuration employed in our experiments.

\begin{table}[!htbp]
\caption{Hyperparameter configuration\label{table:hyperparams}}
\begin{tabular}{lccc}
\hline
\textbf{Parameter} & \multicolumn{3}{l}{\textbf{Values}} \\ \hline
Learning Rate & \multicolumn{3}{c}{1e-4} \\
Batch Size & \multicolumn{3}{c}{64} \\
Number of GNN layers & \multicolumn{3}{c}{3} \\ 
Node Embedding Dimension & \multicolumn{3}{c}{768} \\
Output dimension of GNN & \multicolumn{3}{c}{64} \\ 
Final Embedding Dimension & \multicolumn{3}{c}{64} \\\hline
\end{tabular}
\end{table}
\input{src/table}

\subsection{Evaluation}\label{sec:eval}

\subsubsection{Impact of number of tweets \textbf{(RQ$_1$)}} In this study, we explored the impact of the number of tweets per user per period as it is a crucial parameter, that determines the quantity of data available for the model in the real world. We imposed a fixed number for each experiment, ranging from 1 to 5. The results, presented in Table~\ref{table:vary_tweet_f_4_p_12}, demonstrate that our proposed model \Name outperformed all baseline models when tested with varying levels of content resources. Also, when the quantity of data available to the model increased, the models' performance also improved. This highlights the significance of using pretrained NLP models to obtain rich information from the tweets as node features, as well as the importance of having ample data from users to accurately detect depression. These results underscore the critical role that the number of tweets per user per period plays in depression detection and that utilizing more data leads to superior performance.

\subsubsection{Impact of Number of Friends \textbf{(RQ$_2$)}}
In this study, we aimed to comprehend the extent to which social connections contribute to the identification of depression by manipulating the number of friends per user per period from 2 to 8. The results, as displayed in Table~\ref{table:vary_friend_t_5_p_12}, show that 
\Name significantly outperformed all the baselines by considering both the structural and temporal aspects of social interactions in depression detection. Additionally, the empirical results highlight the importance of considering social connections as a valuable source of information in depression detection and indicate the potential for enhanced performance through the integration of more information from an individual's social milieu.

\subsubsection{Impact of Interval of Period \textbf{(RQ$_3$)}}
In order to gauge the effect of the temporal resolution on the performance of the models, we conducted a comprehensive evaluation by varying the duration of the observational intervals from 3 to 12 months. This experimental setting aimed to examine the relationship between the frequency of observations and the ability of the models to detect depression. The results, as presented in Table~\ref{table:vary_period_t_5_f_4}, indicate a clear correlation between the reduction in the interval duration and an improvement in the models' performance. This can be attributed to the fact that shorter intervals provide more recent and frequent observations of the individual's linguistic and social patterns, thereby increasing the sensitivity of the model to detect depression. Furthermore, the results demonstrate that \Name is robust in early detection of depression even with longer observational intervals. This can be especially valuable in scenarios where a higher frequency of data collection is challenging.

%% file: src/table.tex
\begin{table*}[]
\caption{Performance comparison by varying the number of tweets in each time period\label{table:vary_tweet_f_4_p_12}}
\bgroup
\resizebox{\linewidth}{!}{%
  \begin{tabular}{llllllll|lllllll|lllllll|}
    &\multicolumn{7}{c}{1 tweet}&\multicolumn{7}{c}{3 tweet}&\multicolumn{7}{c}{5 tweet}\\
    &\multicolumn{3}{c}{Healthy (0)} &  \multicolumn{1}{c}{} & \multicolumn{3}{c}{Depessed (1)}&\multicolumn{3}{c}{Healthy (0)} &  \multicolumn{1}{c}{} & \multicolumn{3}{c}{Depessed (1)}&\multicolumn{3}{c}{Healthy (0)} &  \multicolumn{1}{c}{} & \multicolumn{3}{c}{Depessed (1)}\\
    &P&R&F$_{1}$&AUC&P&R&F$_{1}$&P&R&F$_{1}$&AUC&P&R&F$_{1}$&P&R&F$_{1}$&AUC&P&R&F$_{1}$\\
    GCN&0.83&0.82&0.82&0.76&0.69&0.70&0.69&0.85&0.85&0.85&0.80&0.74&0.74&0.74&0.87&0.88&0.87&0.82&0.79&0.77&0.78\\
    GAT&0.84&0.83&0.83&0.77&0.72&0.71&0.71&0.85&0.85&0.85&0.79&0.74&0.73&0.73&0.88&0.85&0.86&0.82&0.76&0.80&0.78\\
    GraphSAGE&0.82&\textbf{0.86}&0.84&0.77&0.73&0.67&0.70&0.85&\textbf{0.86}&0.85&0.79&0.75&0.73&0.73&0.86&0.88&0.87&0.82&0.79&0.76&0.77\\
    DySAT&0.85&0.84&0.84&0.79&0.72&0.73&0.72&0.89&0.82&0.85&0.82&0.73&0.82&0.76&0.89&0.86&0.87&0.84&0.78&0.82&0.79\\
    
    MentalNet&0.88&0.85&0.86&0.82&0.76&0.79&0.77&0.89&0.85&0.86&0.83&0.76&0.81&0.77&0.88&\textbf{0.89}&0.88&0.84&0.80&0.78&0.78\\
    ContrastEgo&\textbf{0.90}&\textbf{0.86}&\textbf{0.88}&\textbf{0.85}&\textbf{0.77}&\textbf{0.83}&\textbf{0.80}&\textbf{0.91}&\textbf{0.86}&\textbf{0.88}&\textbf{0.86}&\textbf{0.78}&\textbf{0.86}&\textbf{0.81}&\textbf{0.92}&0.88&\textbf{0.90}&\textbf{0.88}&\textbf{0.81}&\textbf{0.88}&\textbf{0.84}\\

 \end{tabular}
}
\egroup
\end{table*}

\begin{table*}[]
\caption{Performance comparison by varying the number of friends in each time period\label{table:vary_friend_t_5_p_12}}
\bgroup
\resizebox{\linewidth}{!}{%
  \begin{tabular}{llllllll|lllllll|lllllll|}
    &\multicolumn{7}{c}{2 friend}&\multicolumn{7}{c}{4 friend}&\multicolumn{7}{c}{8 friend}\\
    &\multicolumn{3}{c}{Healthy (0)} &  \multicolumn{1}{c}{} & \multicolumn{3}{c}{Depessed (1)}&\multicolumn{3}{c}{Healthy (0)} &  \multicolumn{1}{c}{} & \multicolumn{3}{c}{Depessed (1)}&\multicolumn{3}{c}{Healthy (0)} &  \multicolumn{1}{c}{} & \multicolumn{3}{c}{Depessed (1)}\\
    &P&R&F$_{1}$&AUC&P&R&F$_{1}$&P&R&F$_{1}$&AUC&P&R&F$_{1}$&P&R&F$_{1}$&AUC&P&R&F$_{1}$\\
    GCN&0.87&0.86&0.87&0.83&0.77&0.79&0.78&0.87&0.88&0.87&0.82&0.79&0.77&0.78&0.90&0.82&0.85&0.82&0.74&0.82&0.78\\
    GAT&0.87&0.87&0.87&0.82&0.78&0.77&0.77&0.88&0.85&0.86&0.82&0.76&0.80&0.78&0.87&0.87&0.87&0.82&0.78&0.77&0.77\\
    GraphSAGE&0.86&0.87&0.87&0.82&0.78&0.76&0.77&0.86&0.88&0.87&0.82&0.79&0.76&0.77&0.87&0.87&0.87&0.82&0.78&0.76&0.77\\
    DySAT&0.89&0.85&0.86&0.83&0.77&0.82&0.79&0.89&0.86&0.87&0.84&0.78&0.82&0.79&0.89&0.84&0.87&0.84&0.76&0.83&0.79\\
    
    MentalNet&0.88&\textbf{0.88}&0.87&0.83&\textbf{0.79}&0.79&0.78&0.88&\textbf{0.89}&0.88&0.84&0.80&0.78&0.78&0.89&0.89&0.89&0.85&0.82&0.80&0.81\\
    ContrastEgo&\textbf{0.90}&0.87&\textbf{0.88}&\textbf{0.85}&\textbf{0.79}&\textbf{0.83}&\textbf{0.81}&\textbf{0.92}&0.88&\textbf{0.90}&\textbf{0.88}&\textbf{0.81}&\textbf{0.88}&\textbf{0.84}&\textbf{0.92}&\textbf{0.90}&\textbf{0.91}&\textbf{0.88}&\textbf{0.84}&\textbf{0.86}&\textbf{0.85}
 
 \end{tabular}
}
\egroup
\end{table*}

\begin{table*}[]
\caption{Performance comparison by varying the duration of time period\label{table:vary_period_t_5_f_4}}
\bgroup
\resizebox{\linewidth}{!}{%
  \begin{tabular}{llllllll|lllllll|lllllll|}
    &\multicolumn{7}{c}{3 months}&\multicolumn{7}{c}{6 months}&\multicolumn{7}{c}{12 months}\\
    &\multicolumn{3}{c}{Healthy (0)} &  \multicolumn{1}{c}{} & \multicolumn{3}{c}{Depessed (1)}&\multicolumn{3}{c}{Healthy (0)} &  \multicolumn{1}{c}{} & \multicolumn{3}{c}{Depessed (1)}&\multicolumn{3}{c}{Healthy (0)} &  \multicolumn{1}{c}{} & \multicolumn{3}{c}{Depessed (1)}\\
    &P&R&F$_{1}$&AUC&P&R&F$_{1}$&P&R&F$_{1}$&AUC&P&R&F$_{1}$&P&R&F$_{1}$&AUC&P&R&F$_{1}$\\
    GCN&0.89&\textbf{0.89}&0.89&0.85&0.81&0.80&0.81&0.88&0.84&0.86&0.82&0.73&0.80&0.76&0.87&0.88&0.87&0.82&0.78&0.76&0.77\\
    GAT&0.90&0.87&0.88&0.85&0.79&0.82&0.81&0.88&0.85&0.86&0.82&0.74&0.79&0.77&0.88&0.85&0.87&0.82&0.75&0.80&0.77\\
    GraphSAGE&0.89&0.88&0.89&0.85&0.80&0.82&0.81&0.88&0.86&0.87&0.83&0.76&0.79&0.77&0.86&0.88&0.87&0.82&0.79&0.75&0.77\\
    DySAT&0.91&0.87&0.89&0.85&0.79&0.84&0.81&0.88&0.85&0.87&0.82&0.75&0.80&0.77&0.89&0.86&0.87&0.84&0.78&0.82&0.79\\
    MentalNet&0.90&\textbf{0.89}&0.89&0.86&0.81&0.82&0.82&0.88&\textbf{0.90}&\textbf{0.89}&0.84&\textbf{0.81}&0.78&0.79&0.87&\textbf{0.89}&0.88&0.83&0.79&0.77&0.78\\
    ContrastEgo&\textbf{0.93}&\textbf{0.89}&\textbf{0.91}&\textbf{0.89}&\textbf{0.82}&\textbf{0.89}&\textbf{0.85}&\textbf{0.92}&0.87&\textbf{0.89}&\textbf{0.87}&0.79&\textbf{0.86}&\textbf{0.82}&\textbf{0.92}&\textbf{0.88}&\textbf{0.90}&\textbf{0.87}&\textbf{0.80}&\textbf{0.87}&\textbf{0.83}\\
 \end{tabular}
}
\egroup
\end{table*}

%% file: src/ablation.tex
\subsection{Ablation Study}
In this subsection, we conducted a series of ablation experiments to verify the effectiveness of each component used in \Name. The variants of \Name we compare are as follows: 

\begin{itemize}
    \item \textbf{\Name.} This is the complete version of our proposed model, including  heterogeneous graph learning, temporal graph learning, supervised contrastive learning, and graph combination.
    \item \textbf{\Variant{C}.} This is a degraded version with the supervised contrastive learning component removed.
    \item \textbf{\Variant{HO}.} This is the degraded version with the heterogeneous graph learning replaced by homogeneous graph learning.
    \item \textbf{\Variant{H}.} In this degraded version, the heterogeneous graph learning module is removed.
    \item \textbf{\Variant{T}.} This is the degraded version with the transformer encoder module removed.
\end{itemize}

\input{src/ablation_table.tex}

\subsubsection{Heterogeneous Graph Learning} This experiment investigated the effect of removing the heterogeneous graph learning component from \Name, denoted by \Variant{H}. This variant relied on the transformer encoder for depression prediction and disregarded information from social networks. Our results, shown in Table~\ref{ablation:vary_tweet_f_4_p_12}, Table~\ref{ablation:vary_friend_t_5_p_12}, and Table~\ref{ablation:vary_period_t_5_f_4}, demonstrate a significant degradation in performance in all settings without heterogeneous graph learning. This highlights the critical significance of social network information in depression prediction, underscoring the importance of social networks as indicators of an individual's behavior and emotional state. Furthermore, we compared the performance of \Name, to a version using homogeneous graph learning instead, denoted by \Variant{HO}. The results demonstrate that \Name outperformed \Variant{HO}, confirming that capturing heterogeneity within social networks is essential for accurate depression prediction. These results suggest that heterogeneous graph learning plays a crucial role in effectively capturing the structure of the social network, thereby improving performance.

\subsubsection{Temporal Graph Learning} This experiment evaluated the impact of removing the transformer encoder component in regard to capturing temporal dependencies, denoted by \Variant{T}. Our results, displayed in Table~\ref{ablation:vary_tweet_f_4_p_12}, Table~\ref{ablation:vary_friend_t_5_p_12}, and Table~\ref{ablation:vary_period_t_5_f_4}, exhibit that the model's performance was slightly degraded without the transformer encoder component in most settings. This suggests that the temporal aspect of the user data is crucial in determining its contribution to the development of depression, as it may vary. Furthermore, the sequence of behaviors can have utterly distinct interpretations and influence on the development of depression, and, hence, should be considered accordingly.

\subsubsection{Supervised Contrastive Learning} 
\begin{figure*}[ht]
    \begin{minipage}[t]{1\linewidth}
        \centering
        \subfigure[\Name]
        {\includegraphics[width=0.245\linewidth]{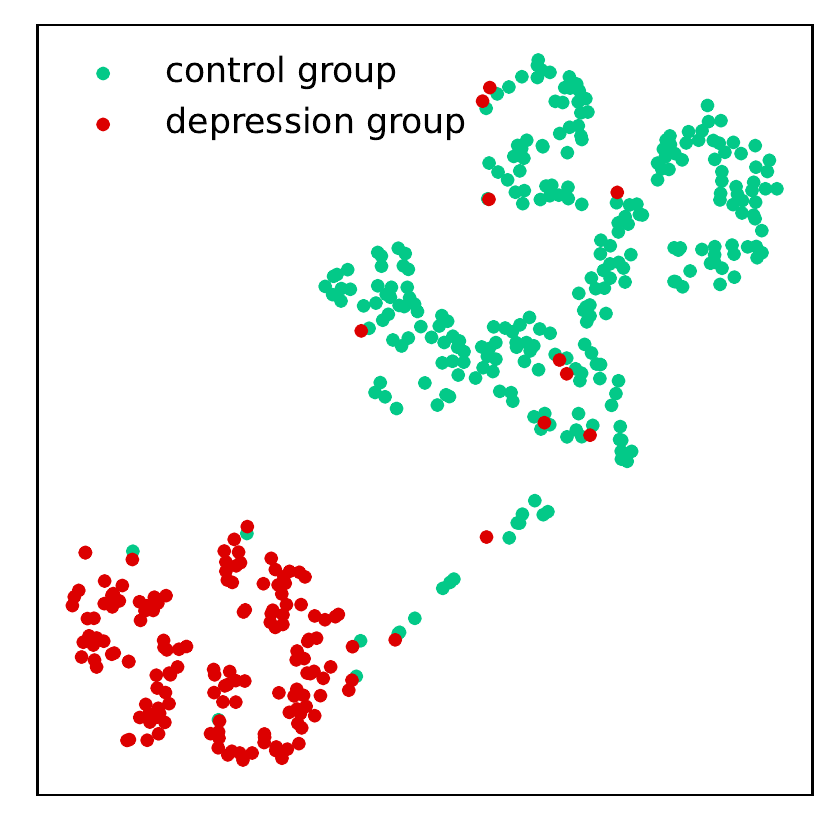}}
        \subfigure[\Variant{C}]
        {\includegraphics[width=0.245\linewidth]{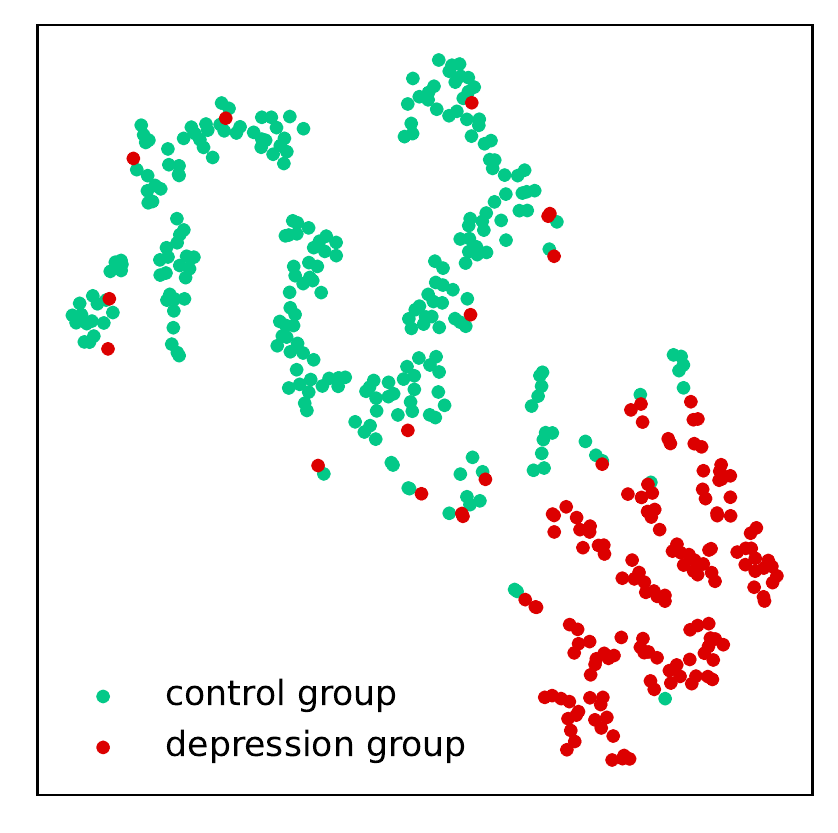}}
        \centering
        \subfigure[\Variant{H}]
        {\includegraphics[width=0.245\linewidth]{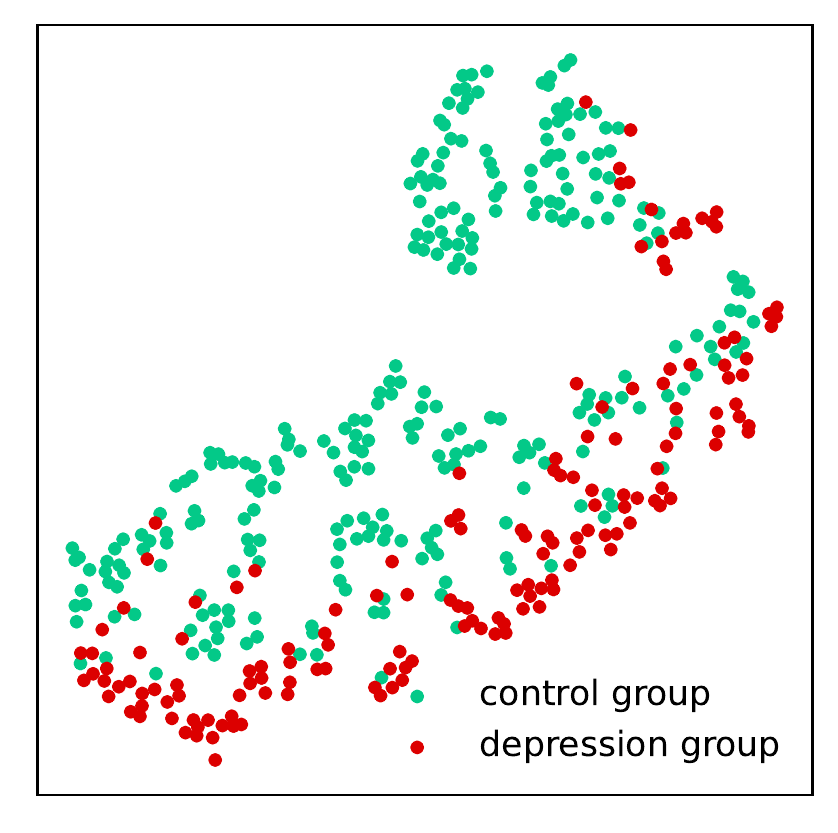}}
        \subfigure[\Variant{T}]
        {\includegraphics[width=0.245\linewidth]{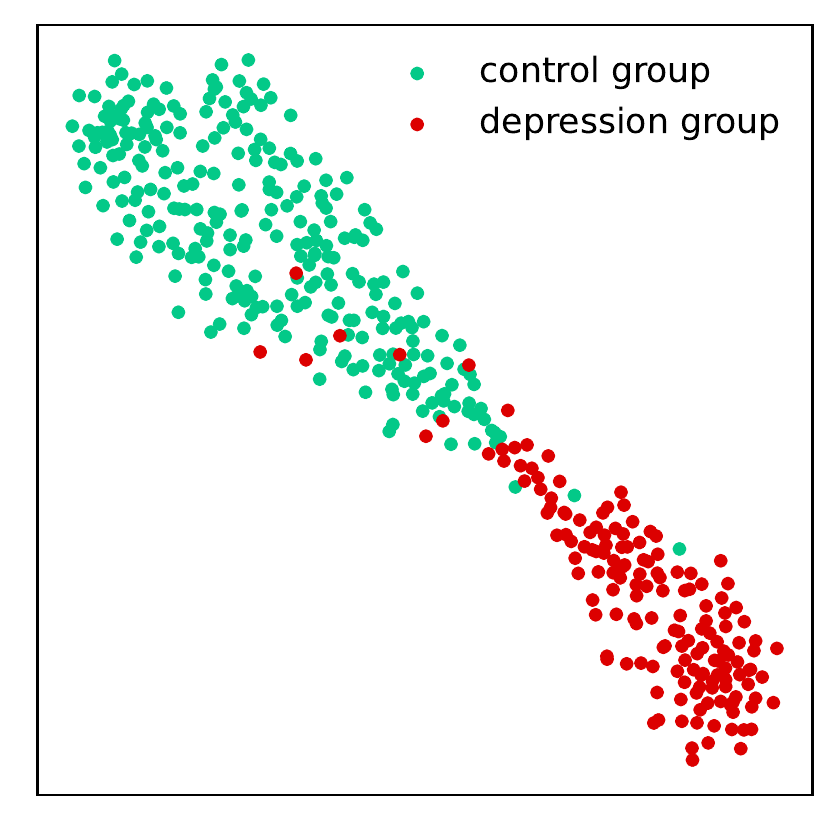}}
        \label{fig:real_time}
    \end{minipage}
    \caption{t-SNE Visualization of User Representations\label{fig:contrastive}}
    \vspace{-1mm}
\end{figure*}

This experiment evaluated the impact of removing the supervised contrastive learning component, denoted by \Variant{C}, on the model's performance. The results shown in Table~\ref{ablation:vary_tweet_f_4_p_12}, Table~\ref{ablation:vary_friend_t_5_p_12}, and Table~\ref{ablation:vary_period_t_5_f_4}, exhibit that the model's performance was slightly degraded without contrastive learning. This suggests that using contrastive learning effectively enhanced the representation of individuals with depression and contributes to the overall performance of the model. The results are further visualized through Figure~\ref{fig:contrastive}, which serves to illustrate the differences between representation of individuals with and without depression in the embedding space.

\subsubsection{Graph Combination} The study compared the efficiency of the proposed graph combination technique to a variant without the graph combination component, denoted by \Variant{GC}. The processing time of \Name was significantly faster with an average of 2.15 seconds per epoch compared to the average of 19.92 seconds per epoch for \Variant{GC}. This is because the proposed technique processes multiple graph snapshots from the same user and graphs from different users together to achieve highly parallelizable processing.

%% file: src/ablation_table.tex
\begin{table*}[]
\caption{num\_friend: 4, period\_in\_months: 12\label{ablation:vary_tweet_f_4_p_12}}
\bgroup
\resizebox{\linewidth}{!}{%
  \begin{tabular}{llllllll|lllllll|lllllll|}
    &\multicolumn{7}{c}{1 tweet}&\multicolumn{7}{c}{3 tweet}&\multicolumn{7}{c}{5 tweet}\\
    &\multicolumn{3}{c}{Healthy (0)} &  \multicolumn{1}{c}{} & \multicolumn{3}{c}{Depessed (1)}&\multicolumn{3}{c}{Healthy (0)} &  \multicolumn{1}{c}{} & \multicolumn{3}{c}{Depessed (1)}&\multicolumn{3}{c}{Healthy (0)} &  \multicolumn{1}{c}{} & \multicolumn{3}{c}{Depessed (1)}\\
    &P&R&F$_{1}$&AUC&P&R&F$_{1}$&P&R&F$_{1}$&AUC&P&R&F$_{1}$&P&R&F$_{1}$&AUC&P&R&F$_{1}$\\
    ContrastEgo&\textbf{0.90}&\textbf{0.86}&\textbf{0.88}&\textbf{0.84}&\textbf{0.77}&\textbf{0.83}&\textbf{0.80}&\textbf{0.91}&\textbf{0.86}&\textbf{0.88}&\textbf{0.86}&\textbf{0.77}&\textbf{0.85}&\textbf{0.81}&\textbf{0.92}&\textbf{0.88}&\textbf{0.90}&\textbf{0.87}&\textbf{0.80}&\textbf{0.87}&\textbf{0.83}\\
    ContrastEgo-C&\textbf{0.90}&0.84&0.87&0.83&0.74&\textbf{0.83}&0.78&0.90&0.83&0.87&0.84&0.74&0.84&0.79&0.91&0.86&0.88&0.85&0.78&0.84&0.81\\
    ContrastEgo-H&0.84&0.76&0.80&0.75&0.63&0.74&0.68&0.85&0.81&0.83&0.78&0.69&0.74&0.71&0.87&0.82&0.85&0.80&0.72&0.78&0.75\\
    ContrastEgo-T&0.89&\textbf{0.86}&\textbf{0.88}&\textbf{0.84}&\textbf{0.77}&0.81&0.79&0.90&0.85&\textbf{0.88}&0.84&0.76&0.84&0.80&0.89&0.87&0.88&0.85&0.79&0.82&0.80\\
    
    ContrastEgo-HO&0.87&\textbf{0.86}&0.87&0.82&0.76&0.78&0.77&0.88&0.85&0.87&0.82&0.75&0.80&0.77&0.90&0.85&0.87&0.84&0.76&0.82&0.79
 \end{tabular}
}
\egroup
\end{table*}

\begin{table*}[]
\caption{num\_tweet\_per\_period: 5, period\_in\_months: 12\label{ablation:vary_friend_t_5_p_12}}
\bgroup
\resizebox{\linewidth}{!}{%
  \begin{tabular}{llllllll|lllllll|lllllll|}
    &\multicolumn{7}{c}{2 friend}&\multicolumn{7}{c}{4 friend}&\multicolumn{7}{c}{8 friend}\\
    &\multicolumn{3}{c}{Healthy (0)} &  \multicolumn{1}{c}{} & \multicolumn{3}{c}{Depessed (1)}&\multicolumn{3}{c}{Healthy (0)} &  \multicolumn{1}{c}{} & \multicolumn{3}{c}{Depessed (1)}&\multicolumn{3}{c}{Healthy (0)} &  \multicolumn{1}{c}{} & \multicolumn{3}{c}{Depessed (1)}\\
    &P&R&F$_{1}$&AUC&P&R&F$_{1}$&P&R&F$_{1}$&AUC&P&R&F$_{1}$&P&R&F$_{1}$&AUC&P&R&F$_{1}$\\
    ContrastEgo&\textbf{0.90}&0.87&\textbf{0.89}&\textbf{0.85}&0.79&\textbf{0.82}&\textbf{0.81}&\textbf{0.92}&\textbf{0.88}&\textbf{0.90}&\textbf{0.87}&\textbf{0.80}&\textbf{0.87}&\textbf{0.83}&\textbf{0.92}&0.90&\textbf{0.91}&\textbf{0.88}&0.83&\textbf{0.85}&\textbf{0.84}\\
    ContrastEgo-C&0.89&0.87&0.88&0.84&0.78&0.80&0.79&0.91&0.86&0.88&0.85&0.78&0.84&0.81&0.91&0.88&0.90&0.86&0.80&\textbf{0.85}&0.82\\
    ContrastEgo-H&0.88&0.80&0.84&0.80&0.70&0.80&0.75&0.87&0.82&0.85&0.80&0.72&0.78&0.75&0.88&0.83&0.86&0.82&0.73&0.80&0.76\\
    ContrastEgo-T&0.87&\textbf{0.91}&\textbf{0.89}&0.84&\textbf{0.83}&0.76&0.79&0.89&0.87&0.88&0.85&0.79&0.82&0.80&0.90&\textbf{0.91}&0.90&0.86&\textbf{0.84}&0.82&0.83\\
    
    ContrastEgo-HO&0.89&0.86&0.87&0.83&0.77&0.80&0.79&0.90&0.85&0.87&0.84&0.76&0.82&0.79&0.90&0.88&0.89&0.85&0.79&0.82&0.81
 \end{tabular}
}
\egroup
\end{table*}
\begin{table*}[]
\caption{num\_tweet\_per\_period: 5, num\_friend: 4\label{ablation:vary_period_t_5_f_4}}
\bgroup
\resizebox{\linewidth}{!}{%
  \begin{tabular}{llllllll|lllllll|lllllll|}
    &\multicolumn{7}{c}{3 months}&\multicolumn{7}{c}{6 months}&\multicolumn{7}{c}{12 months}\\
    &\multicolumn{3}{c}{Healthy (0)} &  \multicolumn{1}{c}{} & \multicolumn{3}{c}{Depessed (1)}&\multicolumn{3}{c}{Healthy (0)} &  \multicolumn{1}{c}{} & \multicolumn{3}{c}{Depessed (1)}&\multicolumn{3}{c}{Healthy (0)} &  \multicolumn{1}{c}{} & \multicolumn{3}{c}{Depessed (1)}\\
    &P&R&F$_{1}$&AUC&P&R&F$_{1}$&P&R&F$_{1}$&AUC&P&R&F$_{1}$&P&R&F$_{1}$&AUC&P&R&F$_{1}$\\
    ContrastEgo&\textbf{0.93}&0.89&0.91&\textbf{0.89}&0.82&\textbf{0.89}&0.85&\textbf{0.92}&0.87&\textbf{0.89}&\textbf{0.87}&0.79&\textbf{0.86}&\textbf{0.82}&\textbf{0.92}&\textbf{0.88}&\textbf{0.90}&\textbf{0.87}&\textbf{0.80}&\textbf{0.87}&\textbf{0.83}\\
    ContrastEgo-C&0.92&0.89&0.90&0.87&0.81&0.86&0.83&0.91&0.88&\textbf{0.89}&0.86&0.79&0.84&\textbf{0.82}&0.91&0.86&0.88&0.85&0.78&0.84&0.81\\
    ContrastEgo-H&0.90&0.88&0.89&0.85&0.80&0.82&0.81&0.88&0.87&0.88&0.83&0.77&0.78&0.78&0.87&0.82&0.85&0.80&0.72&0.78&0.75\\
    
    ContrastEgo-T&0.92&\textbf{0.92}&\textbf{0.92}&\textbf{0.89}&\textbf{0.86}&0.87&\textbf{0.86}&0.90&\textbf{0.89}&\textbf{0.89}&0.85&\textbf{0.80}&0.82&0.81&0.89&0.87&0.88&0.85&0.79&0.82&0.80\\
    
    ContrarstEgo-HO&0.92&0.89&0.91&0.88&0.82&0.87&0.84&0.91&0.86&0.88&0.85&0.77&0.84&0.80&0.90&0.85&0.87&0.84&0.76&0.82&0.79
 \end{tabular}
}
\egroup
\end{table*}

%% file: src/relatedwork.tex
\section{Related Work}\label{sec:related_work}
The utilization of online social media as a source of information for the study of mental health, specifically in the realm of depression detection, has been the subject of considerable scholarly attention in recent years. In particular, NLP techniques and feature engineering on social media data have been the focus of a plethora of studies aimed at detecting depression in an early stage. 

% Machine-Learning-Based Depression Detection
\vspace{1mm}
\noindent\textbf{Machine Learning-Based Depression Detection.} 
Traditional machine learning-based approaches for depression detection focus on identifying relevant features from data and using them as input to machine learning models. Different feature extraction methods have been utilized, such as term-document matrix (TDM) and term frequency-inverse document frequency (TF-IDF)~\cite{Chatterjee_2021}, a bag-of-words model ~\cite{8389299}, and linguistic inquiry and word count (LIWC) for categorizing words based on linguistic properties ~\cite{preotiuc-pietro-etal-2015-role, DeChoudhury_Gamon_Counts_Horvitz_2021}. Several studies have explored the use of multiple feature engineering techniques. Tadesse~\etal~\cite{8681445} combined LIWC, Latent Dirichlet Allocation (LDA), and n-grams for feature extraction and evaluated with different classifiers. Shen\etal~\cite{ijcai2017p536} extracted several depression-related features using LIWC, and LDA, and applied multimodal dictionary learning for sparse user representations.

% Deep learning
\vspace{1mm}
\noindent\textbf{Deep-Learning-Based Depression Detection.} Deep learning methods have been increasingly used in recent years for the prediction of depression from social media data. For instance, Ghosh\etal~\cite{9447025} employed an LSTM network based on a comprehensive set of user features and depression-related n-grams. Shah\etal~\cite{9231008} applied a bidirectional LSTM with various word embedding techniques. Yates~\cite{yates-etal-2017-depression} applied a CNN to each post, all of which are later merged with another convolutional layer to generate a user representation for prediction. Zogan\etal~\cite{10.1145/3404835.3462938} presented a model that integrates a CNN with a stacked bidirectional GRU (BiGRU) with Attention, drawing upon a fusion of user behavior and posting history features, where extraneous content is disregarded, to extract various features through LDA such as social network, emotions, depression domain-specific and topic features. Orabi \etal~\cite{husseini-orabi-etal-2018-deep} explored the performance of various CNN and RNN models using a combination of word embedding techniques for embedding refinement. \cite{FIGUEREDO2022100225} combined a pre-trained NLP model with CNNs using both early and late fusion approaches. Ophir\etal~\cite{Ophir2020} predicted suicide risk from Facebook posts by considering multiple mediating factors, employing a hierarchical subnetwork architecture with a shared layer of fully connected nodes to mitigate the risk of overfitting. Pirayesh\etal~\cite{10.1145/3459637.3482366} proposed a method for obtaining user embeddings based on aggregated social media posts using a triplet network, after which 
dynamic mean shift pruning is used to identify homogeneous friends from social circles for prediction along with the target user. Mihov\etal~\cite{mentalnet} constructed a heterogeneous graph capturing various types of interactions between users and friends with a sort pooling layer for graph classification.

\vspace{1mm}
\noindent\textbf{Temporal/Dynamic Graph Learning.} Temporal graph and dynamic graph learning are emerging fields that study the evolution and changes of graphs over time, and how to model these changes to capture the dynamic nature of real-world events. Pareja\etal~\cite{evolvegcn} introduces an RNN that incorporates the GCN weights and node embeddings to capture the temporal dynamics of graph data. Sankar\etal~\cite{dysat} presented an approach that leverages the use of structural and temporal self-attention mechanisms for learning node representations over time. This method involves processing each graph snapshot through a structural self-attention network, adding positional embeddings, and finally learning the node representations through the temporal self-attention mechanism.

%DyFromer aggregated multiple dynamic graphs into a big temporal graph and encoded temporal encoding (edges between nodes at a specific time-step) and Spatial distance 

%% file: src/conclusion.tex
\section{Conclusion}\label{sec:conclusion}
In conclusion, current depression detection methods have several limitations, particularly their dependence on feature engineering and inadequate consideration of time-varying factors. To tackle these limitations, we have proposed an early depression detection framework, ContrastEgo. Our framework effectively learns multi-scale user latent representations by capturing the topological dynamics of users' social interactions through supervised contrastive learning. Our experiments on Twitter data reveal the importance of taking into account the influence of social circles and time-varying patterns in depression detection. The results demonstrate the superiority of ContrastEgo over existing state-of-the-art GNN-based methods. With up to 5\% improvement in F1 score and 4\% improvement in AUROC compared to the baseline methods, our framework provides a promising solution for early detection of depression.